\documentclass[10pt,twocolumn,letterpaper]{article}

\usepackage{cvpr}              

\usepackage{graphicx}
\usepackage{amsmath}
\usepackage{amssymb}
\usepackage{booktabs}
\usepackage{multirow}
\usepackage{threeparttable}
\usepackage{indentfirst} 
\usepackage{makecell}
\usepackage[lined,boxed,commentsnumbered,ruled]{algorithm2e}
\usepackage{float}
\usepackage[numbers,sort&compress]{natbib}
\usepackage{bm}

\usepackage[pagebackref,breaklinks,colorlinks]{hyperref}

\usepackage[capitalize]{cleveref}
\crefname{section}{Sec.}{Secs.}
\Crefname{section}{Section}{Sections}
\Crefname{table}{Table}{Tables}
\crefname{table}{Tab.}{Tabs.}

\def\cvprPaperID{2936} 
\def\confName{CVPR}
\def\confYear{2024}

\begin{document}

\title{MOSE: Boosting Vision-based Roadside 3D Object Detection with Scene Cues}

\author{Xiahan Chen, Mingjian Chen, Sanli Tang\thanks{Corresponding author}, Yi Niu, Jiang Zhu\\
Hikvision Research Institute \\
{\tt\small \{chenxiahan, chenmingjian, tangsanli, niuyi, zhujiang.hri \} @hikvision.com}
}
\maketitle

\begin{abstract}
\vspace{-0.3cm}
3D object detection based on roadside cameras is an additional way for autonomous driving to alleviate the challenges of occlusion and short perception range from vehicle cameras. Previous methods for roadside 3D object detection mainly focus on modeling the depth or height of objects, neglecting the stationary of cameras and the characteristic of inter-frame consistency. In this work, we propose a novel framework, namely MOSE, for MOnocular 3D object detection with Scene cuEs. The scene cues are the frame-invariant scene-specific features, which are crucial for object localization and can be intuitively regarded as the height between the surface of the real road and the virtual ground plane. 
In the proposed framework, a scene cue bank is designed to aggregate scene cues from multiple frames of the same scene with a carefully designed extrinsic augmentation strategy. Then, a transformer-based decoder lifts the aggregated scene cues as well as the 3D position embeddings for 3D object location, which boosts generalization ability in heterologous scenes. The extensive experiment results on two public benchmarks demonstrate the state-of-the-art performance of the proposed method, which surpasses the existing methods by a large margin.
\end{abstract}

\vspace{-0.6cm}
\section{Introduction}
\vspace{-0.1cm}
\label{sec:intro}

Camera-based 3D object detection technique draws wide attention for autonomous driving because of its low-cost, dense semantic information, and high resolution properties \cite{MonoGround, CaDDN}. However, the ego-vehicle cameras often suffer from the occlusion and poor perception range, increasing the risk of catastrophic damage \cite{BEVheight}. To address these problems, a promising way is to apply extra 3D object detection from the higher-mounted infrastructure cameras\cite{CalibFree, Colla}.

\begin{figure}
    \centering
    \includegraphics[width=0.47\textwidth]{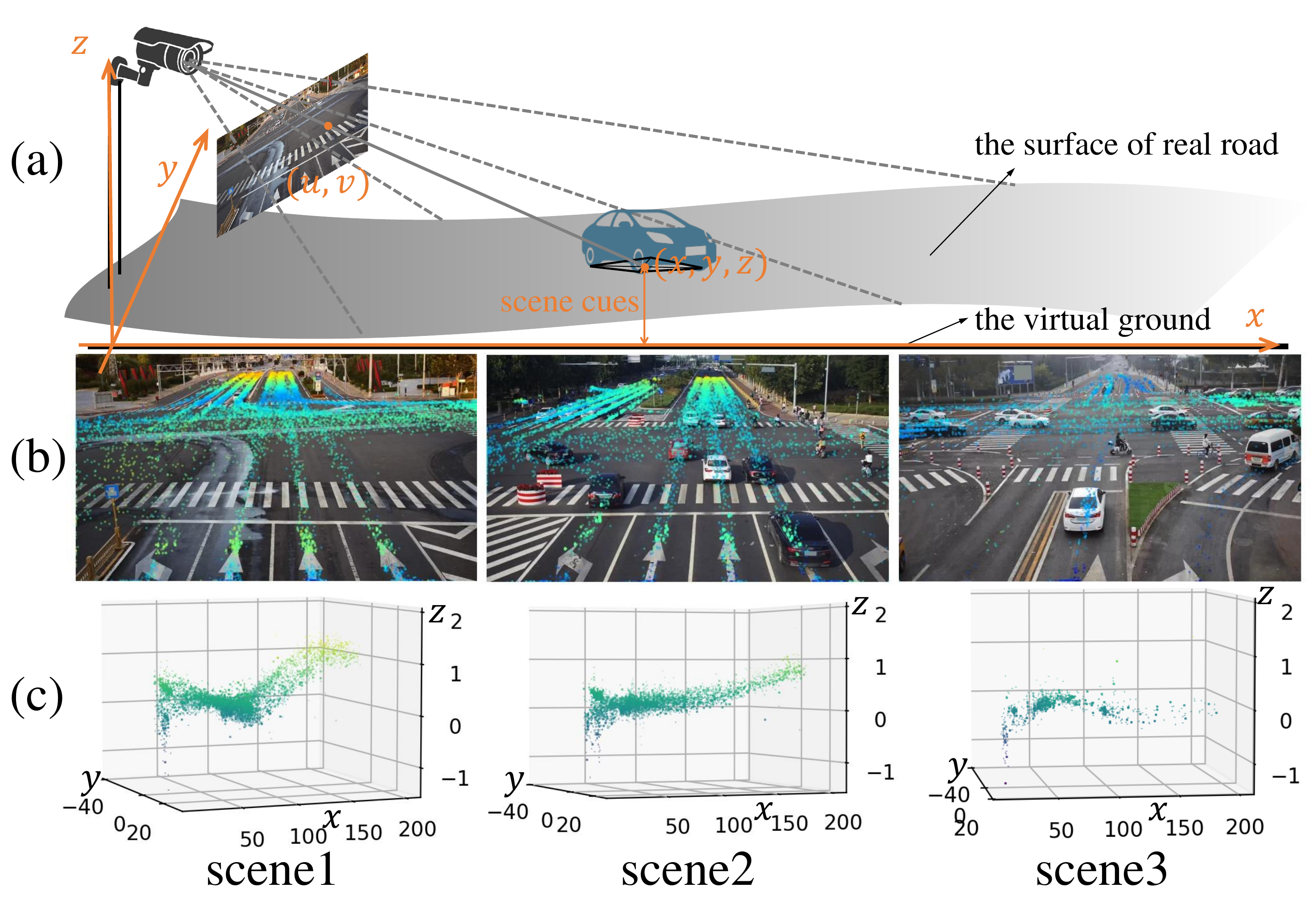}
    \vspace{-0.4cm} 
    \caption{Examples of roadside datasets. (a) Simulation of the roadside camera characteristic. (b) The relative height of all objects on the ground in a roadside scene, which is always smooth and unaltered in the same location. (c)Location distribution of the cars' bottom centers in the ground coordinate.}
    \label{fig:dataset}
    \vspace{-0.5cm} 
\end{figure}

Different from detecting 3D objects on ego-vehicle cameras, the extrinsic disturbance, \emph{e.g.} the various rotations and translations relative to the ground, is common for  roadside camera deployment. Recent researchers \cite{BEVheight,BEVHeight++,MonoGAE} struggle with the extrinsic disturbance. But the performance is still limited. The essence and challenges of 3D detection are lifting 2D  to 3D locations. The poor generalization of predicting the objects' height or depth on the \emph{heterologous} scenes, \emph{i.e.}, images sampled from cameras with different intrinsic, extrinsic parameters and views, will subsequently confuse 2D to 3D projector resulting in low recalls and inaccurate localization. To tackle this problem, some researchers apply data augmentation on  extrinsic parameters \cite{BEVLaneDet} or embed them as features for 3D predictions \cite{BEVheight}. However, those strategies are general without considering the characteristics  of roadside cameras.

Notice that the roadside cameras lies in that they are usually fixed installations. As shown in Fig.\ref{fig:dataset}(a), the frame-invariant, object-invariant and scene-specific features, namely the \emph{scene cues}, are crucial for object localization. Illustratively, the height between the surface of the real road and the virtual ground defined by the extrinsic parameters, are fixed \emph{w.r.t.} the ground. When the object-invariant height is precisely predicted, the BEV location of the object on it can be directly calculated through some geometric deductions. Moreover, there is a unique correspondence between \emph{scene cures} and image pixels, which is an intrinsic characteristic and remains invariant among different frames and objects within a roadside scene. Fig.\ref{fig:dataset}(b) schematically shows the \emph{scene cues} by projecting them onto the images and accumulating them according to the 3D annotations of vehicles from different frames. In Fig.\ref{fig:dataset}(c), the \emph{scene cues} are directly visualized from 3D coordinates defined by the virtual ground. Both of Fig.\ref{fig:dataset}(b) and Fig.\ref{fig:dataset}(c) show the correspondence between \emph{scene cues} and image pixels

In this paper, we propose a novel 3D object detection framework for roadside cameras. A 2D detector is first applied to capture the semantic and pixel coordinates of objects from a single image, which ensures high recall and provides 2D object proposals. Then, based on shared image features, a scene cue bank is established by aggregating and decoupling from global image features \emph{w.r.t.} object proposals. The frame-specific features of the current frame are further fused with the cached \emph{scene cues} from the same scene. A 3D head based on deformable transformer \cite{DeformableDETR} is applied by referring the object proposals, the fused \emph{scene cues}, and a 3D position embedding to capture 3D bounding boxes of objects. Notice that the unique \emph{scene cues} are essentially defined \emph{w.r.t.} the same camera parameters of a scene, thus a hybrid training strategy is also proposed to disambiguate different camera parameter augmentation for the same scene. For model testing, we first update the scene cue bank from multiple frames to retrieve relative robust \emph{scene cues}, and then conduct the 2D and 3D detection for the target frame.

We summarized the main contributions as follows:
\begin{itemize}
\setlength{\itemsep}{0pt}
\setlength{\parsep}{1pt}
\setlength{\parskip}{1pt}

\item We propose roadside \textbf{MO}nocluar 3D detection with \textbf{S}cene cu\textbf{E}s (MOSE), a general and robust vision-based roadside 3D object detection framework that lifts 2D to 3D by aggregating and decoupling the 2D object proposals, the fused \emph{scene cues}, and the 3D position embedding.
\item We explore a scene-specific,  frame-invariant and object-invariant feature, namely \emph{scene cues}, which are revealed crucial for object localization. Furthermore, scene-based camera parameters augmentation is designed to disambiguate different camera parameters for the same scene and increase scene diversity.
\item The expansive experiments show the proposed method achieves  state-of-the-art performance in the public Rope3D and DAIR-V2X datasets. Besides, quantitative and qualitative results reveal that our method is more general to heterologous scenes with different camera intrinsic and extrinsic parameters.

\end{itemize}

\vspace{-0.3cm}
\section{Related Works}
\vspace{-0.1cm}
\label{sec:related}
\noindent \textbf{Monocular 3D Object Detection.} Monocular 3D object detection aims to obtain the 3D location and dimension of interested objects from a single camera image. Existing monocular 3D object detection methods mainly follow the 2D-3D detection pipeline, lifting 2D detection results to the 3D bounding box or adhering to the 2D detection network \cite{MonoDETR}. M3D-RPN \cite{M3D-RPN} is a pioneer work that designs depth-aware convolution to capture 3D spatial features. SMOKE \cite{SMOKE} and FCOS3D \cite{FCOS3D} proposed to detect 3D bounding boxes based on the 2D detection method CenterNet \cite{CenterNet} and FCOS \cite{FCOS} with extra 3D detection head. MonoPair \cite{MonoPair} excavates useful spatial information by predicting the distance of close-by objects.  MonoDLE \cite{MonoDLE} analyses the localization error of monocular 3D detection and proposed 3D IOU loss and distant objects ignoring strategy to tackle the bottleneck. MonoDTR \cite{MonoDTR} introduces a transformer to capture depth and visual features with depth supervision. Recently, MonoDETR \cite{MonoDETR} introduces DETR \cite{DETR} architecture for monocular 3D  detection with depth-guided transformer. The aforementioned methods focus on autonomous driving scenarios, where perception range is limited due to severe occlusion. These methods applied to roadside scenarios will suffer from performance degeneration since roadside cameras can capture long-range objects and own diversified camera parameters. 

\noindent \textbf{Roadside 3D Object Detection}.
Since roadside cameras with higher mounting positions can provide long-range perception for autonomous driving vehicles and intelligent transport systems,  a few works have explored the 3D perception for roadside cameras recently.  CenterLoc3D \cite{CenterLoc3D} and YOLOv7-3D \cite{YOLOv7-3D} simply extends 2D detection method with 3D bounding box supervision. BEVHeight \cite{BEVheight} introduces height to ground prediction instead of depth prediction to transform the 2D image to bird-eye-view (BEV) space. Furthermore, BEVHeight++ \cite{BEVHeight++} improves BEVHeight by incorporating both depth-based and height-based features to be compatible with both autonomous driving and roadside scenarios. 
Although the perception range of BEV-based methods can be enlarged, the methods must make a compromise between the performance and computation cost.  Our method lifts 2D object with fused \emph{scene cues} without any perception range restriction, meeting the long-range perception characteristic of roadside perception.

\noindent \textbf{Temporal Modeling for 3D Object Detection}. 
It's intuitive to model consecutive frames for capturing more geometric clues. Existing works have made great efforts on temporal modeling for autonomous driving. BEVFormer \cite{BEVFormer} utilizes previous frame features by spatial cross-attention to enhance multi-view 3D detection. BEVDet4D \cite{BEVDet4D} introduces temporal information into BEV-based 3D detection by concatenating adjacent BEV features. PETRv2 \cite{PETRv2} extends PETR \cite{PETR} by fusing the position embedding and visual feature of two frames. SOLOFusion \cite{SOLOFusion} further incorporates more than 2 frames to fuse long-time features. StreamPETR \cite{StreamPETR} proposed to reduce computation cost via propagating object queries frame by frame. Different from the aforementioned methods modeling consecutive frames, our proposed method models the \emph{scene cue}, which is time-invariant for a specific scene. The scene cue is established by storing the relative height of the same scene frames, without requiring consecutive frames. It's worth noting that the above temporal modeling methods can not process non-consecutive frames, \emph{i.e.,} these methods can not work on Rope3D and DAIR-V2X datasets.
\begin{figure*}
    \centering
    \includegraphics[width=1.0\textwidth]{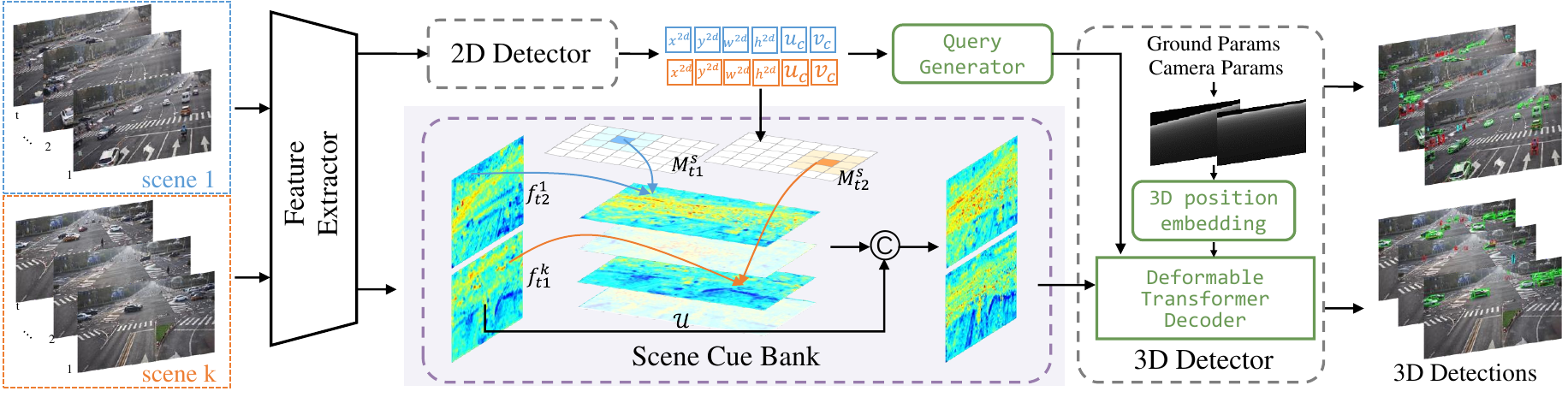}
    \vspace{-0.7cm}
    \caption{Overview framework of our method. A 2D detector is equipped to acquire objects' 2D position information in the current input frame. Then a scene cue bank is proposed to aggregate and decouple object features from image features by 2D object proposals. Finally, a 3D head based on deformable transformer decoders is adopted to infer 3D object bounding boxes.}
    \label{fig:overview}
    \vspace{-0.5cm}
\end{figure*}

\vspace{-0.2cm}
\section{Method}
\vspace{-0.1cm}
\label{sec:method}
\subsection{Preliminary}
\vspace{-0.1cm}
For the 3D object detection task of roadside cameras, the detector takes the image $I_i^t \in \mathbb{R}^{H\times W\times 3}$ as input and predicts the 3D bounding boxes as well as the categories of the interested objects. Formally, the bounding boxes can be denoted as $\mathcal{B}=\{B_i\}_{i=1}^n$ with $B_i=(x_i,y_i,z_i,l_i,w_i,h_i,\theta_i)$, where $(x,y,z)$ is the location of the box. $(l, w, h)$ is the length, width, and height of the box, respectively. $\theta$ denotes the yaw angle, \emph{i.e.}, the orientation of the object \emph{w.r.t.} one pre-defined axis. We use $c_i$ to represent the category of the object with box $B_i$.

The camera parameters, \emph{e.g.}, the intrinsic and extrinsic parameters, can be obtained by careful calibration, which are known for both model training and deployment. By utilizing the extrinsic parameters, a ground coordinate system is established for each camera, denoting as $\mathcal{O}_g$ as in Fig. \ref{fig:2D_to_3D}(a). In practice, the detector $\mathcal{F}$ should predict the 3D bounding boxes \emph{w.r.t.} $\mathcal{O}_g$ from the images as well as the given camera parameters.


\vspace{-0.1cm}
\subsection{Detection from Roadside Camera}
\vspace{-0.1cm}

The essential difference in monocular 3D object detection between vehicle-mounted and roadside cameras is reflected in two aspects: the relative motion of scenes and the disturbance of camera parameters. There are always various types and installation poses (extrinsic) of roadside cameras (intrinsic), which affect 2D to 3D or BEV projection and increase the challenge of generalization ability. Fortunately, the roadside cameras are usually fixed installed, making it possible to enhance the geometric feature learning from multiple frames.

Following the routine of predicting the object heights \cite{BEVheight} to lift from 2D to 3D coordinates, the precision of the predicted heights is essential. As Fig.\ref{fig:2D_to_3D}(b) illustrates, pimping height error will induce large location error, e.g., a height error of 0.5 meters results in a location error of 15 meters when an object is 200 meters away from the camera. 

Instead of relying on the object heights, a more promising way is to predict the distance between the bottom surface of objects and the virtual ground, \emph{i.e.}, the $X_gO_gY_g$ plane, as shown in Fig.\ref{fig:2D_to_3D}(a) and in Fig.\ref{fig:dataset}(a). Such distance $h_r$ can be formulated as
\vspace{-8pt}
\begin{equation}
\label{eq:relativeheight}
h_r=z-h/2
\vspace{-8pt}
\end{equation}

which owns two properties: (1) it is irrelevant to different objects yet it can be acquired by features of objects, which is indeed a scene feature w.r.t the $(x_g, y_g)$ on the virtual ground; (2) there is approximately one-to-one mapping between $(x_g, y_g)$ and the pixel coordinates $(u, v)$, which is invariant within all the frames from the same camera. 

Based on the above considerations, the detector should benefit from those object-invariance and time-invariance features, namely the \emph{scene cues}. For the input frame $I$, aggregating scene cues from past frames should further improve the robustness of the localization prediction.

\subsection{Framework} \label{subsec:Framework}
\vspace{-0.1cm}
\noindent \textbf{Overall Architecture}. As illustrated in Fig.\ref{fig:overview}, the proposed framework first extracts image features $\bm f$ from the current input frame $I$. A 2D detector is equipped to locate and classify objects on the image, which provides object proposals as queries ${\bm q}$. A scene cue bank is followed to extract the scene cues ${\bm f}^s$ for the current frame $s$ based on ${\bm f}$ and ${\bm q}$. Then, ${\bm f}^s$ are aggregated with the memorized scene cues $\hat{{\bm f}}^s$ for more precisely predicting the relative height of objects \emph{w.r.t.} the virtual ground, as defined in Eq.\ref{eq:relativeheight}. Finally, a 3D detector based on deformable transformer \cite{DeformableDETR} takes the aggregated scene cues with the object queries ${\bm q}$ to further predict $h_r$, as well as the sizes $(l,w,h)$ and yaw angle $\theta$ of objects. The location of objects can be straightforwardly calculated by referring $h_r$ as well as the camera parameters.

\noindent \textbf{2D Detector.} The 2D detector is revised from FCOS \cite{FCOS} to acquire the 2D bounding boxes and categories. In the meanwhile, an extra branch is added to regress the pixel point $(u_c,v_c)$ of each object, as illustrated in Fig.\ref{fig:overview}, the ground-truth of which can be obtained by projecting the bottom center points of the object cuboid. The pixel point is designed as the reference point to extract and aggregate scene cues because (1) it is invariant to object orientations and camera parameters; (2) it is corresponding to $h_r(x,y)$ on object location $(x,y)$ regardless of the object height.

\noindent \textbf{Scene Cue Bank.} As shown in Fig.\ref{fig:overview}, the scene cue bank is designed to acquire the scene cues of different scenes by aggregating frame image feature of the same scene. For $S$ scenes in total, the bank contains a buffer 
$\mathcal{U}=\{\hat{\bm{f}^s}\}^S_{s=1} \in \mathbb{R}^{S \times \frac{H}{8}\times \frac{W}{8}\times d}$,
which is used to memorize the aggregated scene cue features from the past frames of the same scene. Taking the image features ${\bm f}$ and the $P$ reference points ${(u_c^i,v_c^i)}_{i=1}^P$ from the 2D detector as inputs, the bank first generates a 0-1 mask ${\bm M}^s$ by activating the 8-neighborhood pixels around the $P$ reference points. 
The scene cue features of current frame ${\bm f}^s$ can be extracted by ${\bm f}^s={\bm f}*{\bm M}^s$. Then, the memorized scene cues $\hat{\bm{f}}^s$ belonging to the scene $s$ can be updated by aggregating $\bm{f}^s$ in momentum way:
\vspace{-6pt}
\begin{equation} \label{eq:G in training}
    \hat{\bm{f}}^s_{t} = (1-\lambda)\hat{\bm{f}}^s_{t-1} + {\lambda}{\bm f}^s,
\vspace{-6pt}
\end{equation}
where the subscript $t$ is denoted the $t^{\rm th}$ timestep for clarity. Then the scene cues $\hat{\bm{f}}^s_{t}$ and the image features of the current frame are concatenated as the outputs of scene cue bank.

Strictly speaking, a general "scene" is defined by the same camera parameters including both intrinsic and extrinsic as well as the camera locations. In order to improve the generalization ability \emph{w.r.t.} the camera parameters, the augmentations on the camera parameters are necessary \cite{BEVheight, MonoGAE, BEVHeight++}. However, randomly augmenting the camera parameters disturbs the proposed scene cue bank to aggregate scene cues via multiple frames. Thus, the training strategy of the scene bank is carefully designed, namely scene-based augmentation strategy. As shown in Alg.\ref{training_algorithm}, for each unique scene in the training set, we keep the augmentation strategy on camera parameters unchanged for some training steps until the number of images from the same scene used for training exceeds a certain threshold $\tau$. The memoried scene cues in the bank are eliminated and initialized with scene cues of the first frame when the general "scene" is replaced by a new one, \emph{e.g.}, another augmentation on the camera parameters. In addition, for model inference, the scene cue bank is first updated from multiple frames and then the 2D and 3D detection is conducted for the target frame, the details of which are shown in supplemental materials.

\noindent \textbf{Detecting 3D object based on transformer.} Since the predictions of 2D bounding boxes as well as the bottom center points $(u_c,v_c)$ are well detected and located for objects on images, the 3D detector should benefit from these coordinates as object queries for 3D detection tasks. In the 3D object detector, a query generator takes the 2D boxes and bottom center points as input, which are then embedded by a sine position encoding layer \cite{DETR} and an MLP layer. The embedded object queries are treated as initial queries for the transformer-based decoder.

To enhance the ability of 3D location and generalization on camera parameters, a 3D position embedding layer is applied by incorporating the camera parameters as the position embedding of the transformer. Concretely, given a pixel point $(u,v)$, the depth $d(u,v)$ between camera and the virtual ground through $(u,v)$ can be solved as 
\vspace{-8pt}
\begin{equation}
d(u,v)=\frac{-D}{\frac{A(u-a_x)}{f_x}+\frac{B(v-a_y)}{f_y}+C}.
\vspace{-8pt}
\end{equation}
$(f_x,f_y,a_x,a_y)$ are from the instrinsic matrix $K$. $(A, B, C, D)$ are the parameters of the virtual ground, which can be calculated from the extrinsic matrix $M$. Thus, the 3D position embedding for each pixel $(u,v)$ can be calculated \emph{w.r.t.} $d(u,v)$ by sine position encoding \cite{DETR}.

We follow the standard deformable transformer \cite{DeformableDETR} that takes the aggregated scene cues with the position embedding, object query as the value, and query for the deformable transformer, respectively. An MLP layer is appended after the transformer to output the final predictions $(l,w,h,h_r,\theta)$. 

\begin{figure}
    \centering
    \includegraphics[width=0.5\textwidth]{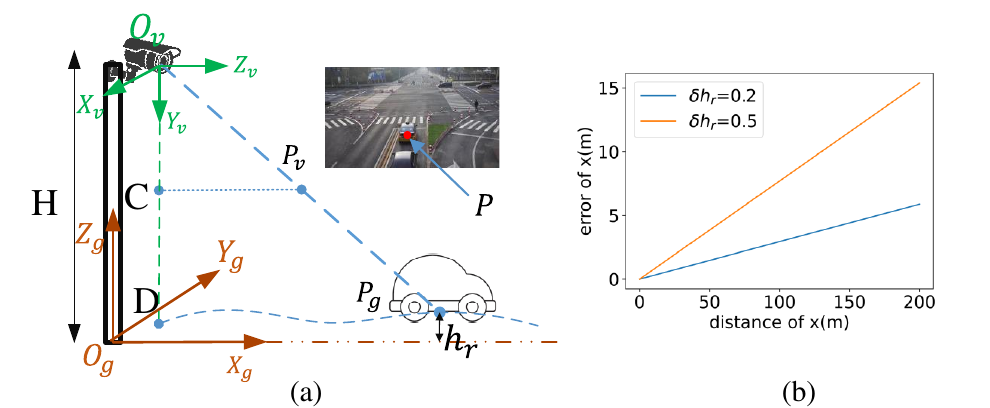}
    \caption{(a)Height-based lifting method. (b) Sensitivity of height error relative to distance. x-axis and y-axis represent the distance and error of x in ground coordinate when camera height $H$ is 7m, and $\delta h_r$ is the prediction error of $h_r$. A tiny error of $h_r$ will lead to serious distance error, especially for long-distance objects.}
    \label{fig:2D_to_3D}
    \vspace{-0.5cm}
\end{figure}

To retrieve 3D coordinates $P_g=(x_g,y_g,z_g)$ from $(h_r,u_c,v_c)$, height-based object lifting is applied, as shown in Fig.\ref{fig:2D_to_3D}(a). The camera coordinate is rotated to obtain the virtual camera coordinate where the plane $X_{v}O_vZ_v$ is parallel to the plane $X_gO_gY_g$. The 3D virtual camera coordinate for bottom center points $(u_c,v_c)$ with unit depth can be calculated as follows:
\vspace{-8pt}
\begin{equation} \label{eq:p_vir}
    P_{v} = M_{\mathrm{cam}}^{v}K^{-1}(u_{c},v_{c},1)^T,
\vspace{-8pt}
\end{equation}
where $M_{\mathrm{cam}}^{v}$ means the transform matrix from camera to the virtual camera coordinate. Then with the similar triangle between $\Delta O_{v}CP_{v}$ and $\Delta O_{v}DP_{g}$, the 3D ground coordinate of bottom center  $(u_c,v_c)$ can be calculated by:
\vspace{-8pt}
\begin{equation} \label{eq:p_ground}
    P_{g} = M_{v}^g\frac{H-h_r}{y_{v}}P_{v}
\vspace{-8pt}
\end{equation}
where $M_{v}^g$ means the transform matrix from the virtual camera coordinate to the ground coordinate and $H$ is the height of $O_{v}$ to the ground plane $X_gO_gZ_g$.

    \vspace{-0.1cm}
\subsection{Optimization}
    \vspace{-0.1cm}
In our proposed method, 2D detector and 3D detector are jointly trained with the shared backbone weight, the total loss consists of two parts: 2D detection loss $L_{2D}$ and 3D detection loss $L_{3D}$. 

\noindent \textbf{2D Loss.} The ${\cal}L_{2D}$ is directly adopted from 2D detector FCOS \cite{FCOS} with extra losses of the bottom center $(u_c,v_c)$ . L1 loss is applied to supervise the bottom center $(u_c,v_c)$. 

\noindent \textbf{3D Loss.} The outputs of our model include the bottom center $(u_c,v_c)$ in pixel space, the height $h_r$ of the object’s bottom center relative to the ground, 3D dimension $(l,w,h)$ of the object, encoded yaw angle $(sin\theta, cos\theta)$. The 3D location $P_{g}$  can be decoded using $(u_c,u_c, h_r)$ by Eq.\ref{eq:p_vir} and Eq.\ref{eq:p_ground}. 

Different from directly measuring the error between the above output and ground truths, we calculate the eight corners of the 3D bounding box using the 3D location $(x,y,z)$ of the bottom center, object 3D dimension $(l,w,h)$, and yaw rotation matrix $R_\theta$ calculated from $\theta$:
\vspace{-8pt}
\begin{equation}
    B_8 = R_\theta (\pm l/2,\pm w/2, \pm h/2)^T + (x,y,z+h/2)^T
\vspace{-8pt}
\end{equation}
The 3D regression loss function can be represented as:
\vspace{-8pt}
\begin{equation}\label{eq:B_8}
    L_{\mathrm{reg}}=||B_8-\bar{B}_8||_1
\vspace{-8pt}
\end{equation}
where $\bar{B}_8$ is the ground truth of the corners.

In order to further improve the detection performance and make the optimization process stable, the model outputs are separated into three parts: location $(x,y,z)$, dimension $(l,w,h)$, and orientation $\theta$. When calculating the loss defined in Eq.\ref{eq:B_8}, two parts of outputs are set to ground truth, and the remainder part is optimized.

Furthermore, we design an extract L1 loss $L_{h_r}$ to supervise the relative height $h_r$, so $L_{3D}$ can be expressed as: 
\vspace{-8pt}
\begin{equation} \label{eq:loss_3d}
    L_{3D} =\lambda_1* \frac{1}{3}\sum_{i=1}^{3}L_{\mathrm{reg}} + \lambda_2* L_{h_r}
\vspace{-8pt}
\end{equation}
where $\lambda_1$ and $\lambda_2$ are balanced parameters.

Finally, the total training loss is described as:
\vspace{-8pt}
\begin{equation} \label{eq:total loss}
    L = L_{2D} + L_{3D}
\vspace{-8pt}
\end{equation}


\begin{algorithm}[t]
\small
\caption{Algorithm for Training 3D Detector} \label{training_algorithm}
\SetAlgoLined
\KwIn{training set $D^{tr}$, scene cue bank $\mathcal{U} \in \mathbb{R}^{S\times\frac{H}{8}\times \frac{W}{8}\times d}$, trained frame number $n$ of each scene}
let e be the current training epoch and set e=0 \\
initialize: $\mathcal{U}$, $n$ \\
\While{$e<T$}{
Sample a batch $D=\{(I_1^1,...I_t^s,... )\}$ from $D^{tr}$ \\
Data augmentation to scene $s$: ${\tilde I}_t^s \leftarrow I_t^s $, $n^s \leftarrow n^s+1$ \\
Extract features ${\bm f}^s$ of current frame, memorized $\hat{\bm{f}}_{t-1}^s$\\
Obtain output $B_t={\cal F}({\tilde I}_t^s, \hat{\bm{f}}_{t-1}^s)$ \\
\eIf{$n^s<\tau$}{
  Update $\hat{\bm{f}}_t^s$ by Eq.\ref{eq:G in training} using ${\bm f}^s$ and $\hat{\bm{f}}_{t-1}^s$
  }
  {
  Initialize $\hat{\bm{f}^s}$, $n^s=0$ \\
  Update date augmentation strategy to scene $i$
  }
Training the detector by minimizing Eq.\ref{eq:total loss} \\
Set $e\leftarrow e+1$\\
}
\end{algorithm}


\begin{table*}[h]
\scriptsize
\centering
\caption{Compared with the start-of-the-art on the DAIR-V2X-I validation set. We report three types of objects, including vehicle(Veh.), pedestrian(Ped.), and cyclist(Cyc.). Each object is categorized into three settings by difficulty introduced in \cite{DAIR-V2X}. L and C mean LiDAR and camera, respectively.}\label{v2x}
\vspace{-0.2cm}
\begin{tabular}{c|c|ccc|ccc|ccc}
\hline 
\multirow{2}{*}{Method} & \multirow{2}{*}{Modal} & \multicolumn{3}{c}{Veh.(IoU=0.5)} & \multicolumn{3}{|c|}{Ped.(IoU=0.25)} & \multicolumn{3}{c}{Cyc.(IoU=0.25)} \\
\cline{3-11}
~ & ~ & Easy & Mod. & Hard & Easy & Mod. & Hard & Easy & Mod. & Hard \\
\hline 
PointPillars \cite{PointPillars} & L & 63.07 & 54.00 & 54.01 & 38.54 & 37.21 & 37.28 & 38.46 & 22.60 & 22.49 \\
SECOND \cite{SECOND} & L & 71.47 & 53.99 & 54.00 & 55.17 & 52.49 & 52.52 & 54.68 & 31.05 & 31.20 \\
\hline 
MVXNET(PE) \cite{MVX-Net} & C+L & 71.04 & 53.72 & 53.76 & 55.83 & 54.46 & 54.40 & 54.05 & 30.79 & 31.07 \\
\hline 
Imvoxelnet \cite{Imvoxelnet} & C & 44.78 & 37.58 & 37.56 & 6.81 & 6.75 & 6.73 & 21.06 & 13.58 & 13.18 \\
MonoDETR \cite{MonoDETR} & C & 58.86 & 51.24 & 51.00 & 16.30 & 15.37 & 15.61 & 37.93 & 34.04 & 33.98 \\
BEVFormer \cite{BEVFormer} & C & 61.37 & 50.73 & 50.73 & 16.89 & 15.82 & 15.95 & 22.16 & 22.13 & 22.06 \\
BEVDepth \cite{BEVDepth} & C & 75.50 & 63.58 & 63.67 & 34.95 & 33.42 & 33.27 & 55.67 & 55.47 & 55.34 \\
BEVHeight \cite{BEVheight} & C & 77.78 & 65.77 & 65.85 & 41.22 & 39.39 & 39.46 & 60.23 & 60.08 & 60.54 \\
BEVHeight++ \cite{BEVHeight++} & C & 79.31 & 68.62 & 68.68 & 42.87 & 40.88 & 41.06 & 60.76 & 60.52 & 61.01 \\
\hline 
MonoGAE \cite{MonoGAE}& C & 84.61 & 75.93 & 74.17 & 25.65 & 24.28 & 24.44 & 44.04 & 47.62 & 46.75 \\
MOSE(ours) & C & \textbf{86.07} & \textbf{81.44} & \textbf{81.47} & 35.73 & 32.15 & 32.12 & 58.38 & 54.12 & 54.25 \\
\hline 
\end{tabular}
\end{table*}

\vspace{-0.3cm}
\section{Experiment}
\vspace{-0.1cm}
\label{sec:experiment}
\subsection{Dataset}
\vspace{-0.1cm}
\noindent \textbf{DAIR-V2X.} DAIR-V2X \cite{DAIR-V2X} is a large-scale and multi-modality vehicle-infrastructure collaborative 3D detection dataset and we focus on the roadside perception subset named DAIR-V2X-I, including 10084 images. DAIR-V2X-I is split into training, validation and testing set by 50\%, 20\% and 30\%, respectively. Up to now, the testing set has not been published, so we evaluate the results on the validation set following previous work \cite{BEVheight}.

\noindent \textbf{Rope3D.} Rope3D \cite{Rope3D} is another large-scale roadside monocular 3D object detection dataset. The dataset consists of over 50k image frames in various scenes at different times, different weather conditions and different densities. We split 70\% of images for training and the remaining 30\% for validation on the original benchmark as reported in \cite{Rope3D}. 
\vspace{-0.5cm}
\subsection{Implementation Details}\label{subsec:Implementation Details}
\vspace{-0.1cm}
We use ResNet101 as the backbone for a fair comparison with state-of-the-art methods \cite{BEVDepth,BEVheight,BEVHeight++,MonoGAE}, and ResNet50 for other analysis and ablation studies. The training process includes two steps: (1) training backbone and 2D head using SGD optimizer with a weight decay of 0.0001 and the initial learning rate of 0.001; (2) jointly training 3D head and 2D head using AdamW optimizer with a weight decay of 0.01 and the initial learning rate of 0.0001. 

In the training process, we conduct data augmentation about camera intrinsic and extrinsic parameters to enrich the camera poses. Specifically,
the image scaling ratio in the camera intrinsic parameters is sampled from the normal distribution of $\mathcal{N} (1, 0.2)$ and clamped with (0.8,0.9). We also apply the random noise with a normal distribution $\mathcal{N}(0, 2)$ along the roll and $\mathcal{N}(0, 0.67)$ along pitch axes for extrinsic parameter augmentation.

Our models except for Sec.\ref{subsec:Results on the original benchmark} are trained on the Rope3D dataset and trained on 8 V100 GPUs for 40 and 24 epochs in two steps, respectively. We focus more on the results of the heterologous validation set as described in Sec.\ref{subsec:Analysis on heterlogous scenes}.

\subsection{Main Results}
\vspace{-0.1cm}
\subsubsection{Results on the original benchmark} \label{subsec:Results on the original benchmark}
\vspace{-0.1cm}
We compare MOSE with other state-of-the-art methods on DAIR-V2X-I and Rope3D. 

\noindent \textbf{Results on the DAIR-V2X-I validation set.} Our proposed method is compared with other state-of-the-art monocular detectors \cite{Imvoxelnet,MonoDETR,BEVFormer,BEVDepth,BEVheight,BEVHeight++,MonoGAE}. For other modality methods \cite{PointPillars,SECOND,MVX-Net}, we directly report their results from \cite{BEVheight} as shown in Table \ref{v2x}. MonoGAE \cite{MonoGAE} is the latest roadside monocular 3D Object detection which achieves high detection accuracy and generalization performance by integrating roadside ground information. Because the split of training and validation setting is not published, we shuffle all images and select 70\% images as the training set and the rest for validation following the instruction in \cite{Rope3D}. For the vehicle category, our method outperforms MonoGAE by 3.12\% and 4.88\% in 'Mod.' and 'Hard' setting, respectively. Although pedestrian and cyclist detection is full of challenges \cite{MonoGAE}, our method still exceeds MonoGAE in all settings. More specific analysis between MOSE and BEVHeight is shown in Sec.\ref{subsec:Analysis on detecting ability} and Fig.\ref{fig:dis_all}. 

\noindent \textbf{Results on the Rope3D validation set.} Following \cite{BEVheight}, we evaluate our method on the Rope3D dataset in car and big vehicle(bus and truck) among BEVDepth \cite{BEVDepth}, BEVHeight \cite{BEVheight}, BEVHeight++ \cite{BEVHeight++}, MonoGAE \cite{MonoGAE}, and other representative methods \cite{M3D-RPN,Kinematic,MonoDLE,MonoFlex,BEVFormer}. As shown in Table \ref{rope3d}, MOSE surpasses MonoGAE by large margins on big vehicle category, \emph{e.g.}, 10.12\% and 8.5\% on metrics of AP and $\mathbf{Rope_{score}}$ \cite{Rope3D}, respectively. For the car category, our method achieves the second-best performance and is better than BEVHeight++.

\begin{table}[h]
\small
\centering
\vspace{-0.1cm}
\caption{Results on the Rope3D validation set in homologous settings. The dataset split and evaluation metrics are following 
\cite{Rope3D}.}\label{rope3d}
\vspace{-0.2cm}
\begin{threeparttable}
\begin{tabular}{c|c|c|c|c}
\hline 
\multirow{2}{*}{Method} & \multicolumn{2}{c}{car} & \multicolumn{2}{|c}{Big Vehicle} \\
\cline{2-5}
~ & AP & Rope & AP & Rope \\
\hline 
M3D-RPN \cite{M3D-RPN} & 54.19 & 62.65 & 33.0 & 44.94 \\
Kinematic3D \cite{Kinematic} & 50.57 & 58.86 & 37.60 & 48.08 \\
MonoDLE \cite{MonoDLE} & 51.70 & 60.36 & 40.34 & 50.07 \\
MonoFlex \cite{MonoFlex} & 60.33 & 66.86 & 37.33 & 47.96 \\
BEVFormer \cite{BEVFormer} & 50.62 & 58.78 & 34.58 & 45.16 \\
BEVDepth \cite{BEVDepth} & 69.63 & 74.70 & 45.02 & 54.64 \\
BEVHeight \cite{BEVheight} & 74.60 & 78.82 & 48.93 & 57.70 \\
BEVHeight++ \cite{BEVHeight++} & 76.12 & 80.91 & 50.11 & 59.92 \\
\hline 
MonoGAE \cite{MonoGAE} & \textbf{80.12} & \textbf{83.76} & \underline{54.62} & \underline{62.37} \\
MOSE(ours) & \underline{76.73} & \underline{80.95} & \textbf{64.74} & \textbf{70.92} \\
\hline 
\end{tabular}
\begin{tablenotes}
    \footnotesize
    \item AP: AP3D$|$R40(IoU=0.5), Rope: $\mathrm{Rope_{score}}$.
\end{tablenotes}
\end{threeparttable}
\end{table}
\vspace{-0.2cm}

\begin{table}[h]
\scriptsize
\setlength\tabcolsep{4pt}
\centering
\vspace{-0.2cm}
\caption{Compared with BEVHeight on the Rope3D validation set in heterologous setting.}\label{table:hete0.5}
\vspace{-0.2cm}
\begin{tabular}{c|c|c|c|c|c|c|c}
\hline 
\multirow{2}{*}{Method} & \multirow{2}{*}{Metric} & \multicolumn{3}{c}{Car(IoU=0.5)} & \multicolumn{3}{|c}{Big Vehicle(IoU=0.5)} \\
\cline{3-8}
~ & ~ & Easy & Mod. & Hard & Easy & Mod. & Hard \\
\hline 
\multirow{2}{*}{BEVheight} & BEV AP & 33.78 & 30.93 & 29.43 & 13.17 & 15.32 & 15.14 \\
 & 3D AP & 20.32 & 18.58 & 18.44 & 7.97 & 9.71 & 9.60 \\
 \hline 
\multirow{2}{*}{MOSE(ours)} & BEV AP & \textbf{45.41} & \textbf{38.33} & \textbf{36.66} & \textbf{19.48} & \textbf{21.94} & \textbf{20.35} \\
 & 3D AP & \textbf{31.10} & \textbf{25.62} & \textbf{25.31} & \textbf{8.97} & \textbf{11.04} & \textbf{10.13} \\
\hline 
\end{tabular}
\vspace{-0.2cm}
\end{table}

\vspace{-0.2cm}
\subsubsection{Analysis on heterlogous scenes.} \label{subsec:Analysis on heterlogous scenes}
\vspace{-0.2cm}

Heterologous scenes are defined as scenes with different camera parameters or scene views from the scenes in the training set. To analyze the generalization ability among heterologous scenes, we compare our method with BEVHeight \cite{BEVheight} with the released codes. Both MOSE and BEVHeight are height-based methods.



In general, each camera is calibrated by different intrinsic parameters and extrinsic parameters for the roadside object detection task, and the fixed background makes it easily overfit in homologous settings. In this subsection, we specially evaluate the generalization ability among the heterologous scenes. To this end, we re-split the Rope3D dataset into the training set and a heterologous validation set. Specifically, 16 scenes are selected for training and the remaining 5 heterologous scenes are used for validation. In the following sections, all results are reported from the heterologous validation set.

The results in Table \ref{table:hete0.5} reveal that although the performance of both MOSE and BEVHeight decreases by margin referring to that in homologous scenes in Table \ref{v2x} and Tabel \ref{rope3d}, MOSE behaves much better than BEVHeight in all metrics. A qualitative comparison is present in supplementary materials. From both quantitative and qualitative analysis, MOSE achieves better generalization ability in heterologous  scenes, which is the major challenge for detecting 3D objects on roadside cameras.

\begin{table*}[h]
\scriptsize
\centering
\vspace{-0.2cm}
\caption{Compared with BEVHeight on the Rope3D validation set.}\label{hete0.1}
\vspace{-0.2cm}
\begin{tabular}{c|c|c|c|c|c|c|c}
\hline 
\multirow{2}{*}{Method} & \multirow{2}{*}{Metric} & \multicolumn{3}{c}{Car(IoU=0.1)} & \multicolumn{3}{|c}{Big Vehicle(IoU=0.1)} \\
\cline{3-8}
~ & ~ & Easy & Mod. & Hard & Easy & Mod. & Hard \\
\hline 
\multirow{2}{*}{BEVheight} & BEV AP & 84.91 & 75.93 & 73.52 & 31.86 & 34.55 & 32.06 \\
 & 3D AP & 83.72 & 74.95 & 70.36 & 31.72 & 34.46 & 31.98 \\
 \hline 
\multirow{2}{*}{MOSE(ours)} & BEV AP & 89.69\textcolor{red}{(+4.78)} & 79.25\textcolor{red}{(+3.32)} & 76.77\textcolor{red}{(+3.25)} & 44.23\textcolor{red}{(+12.38)} & 46.60\textcolor{red}{(+12.04)} & 44.06\textcolor{red}{(+12.00)} \\
 & 3D AP & 88.77\textcolor{red}{(+5.05)} & 76.08\textcolor{red}{(+1.13)} & 73.63\textcolor{red}{(+3.27)} & 41.98\textcolor{red}{(+10.26)} & 45.46\textcolor{red}{(+11.00)} & 41.69\textcolor{red}{(+9.71)} \\
\hline 
\end{tabular}
\end{table*}

\begin{table*}[h]
\scriptsize
\centering
\caption{Ablation study on scene cue bank.}\label{ablation}
\vspace{-0.3cm}
\begin{threeparttable}
\begin{tabular}{c|c|c|c|c|c|c|c|c|c|c|c|c|c}
\hline 
\multirow{2}{*}{SCB} & \multirow{2}{*}{Metric} & \multicolumn{3}{c}{car(IoU=0.5)} & \multicolumn{3}{|c|}{Big Vehicle(IoU=0.5)} & \multicolumn{3}{c}{car(IoU=0.1)} & \multicolumn{3}{|c}{Big Vehicle(IoU=0.1)}\\
\cline{3-14}
~ & ~ & Easy & Mod. & Hard & Easy & Mod. & Hard & Easy & Mod. & Hard & Easy & Mod. & Hard \\
\hline 
\multirow{2}{*}{w/o} & BEV AP & 41.92 & 35.30 & 33.73 & 14.13 & 16.54 & 15.99 & 87.78 & 75.72 & 73.29 & 41.99 & 44.53 & 42.02 \\
 & 3D AP & 28.77 & 24.56 & 23.27 & 7.51 & 9.39 & 8.56 & 84.63 & 74.37 & 70.09 & 39.89 & 43.4 & 39.90 \\
 \hline 
\multirow{2}{*}{w} & BEV AP & 45.41 & 38.33 & 36.66 & 19.48 & 21.94 & 20.35 & 89.69 & 79.25 & 76.77 & 44.23 & 46.60 & 44.06 \\
 & 3D AP & 31.10 & 25.62 & 25.31 & 8.97 & 11.04 & 10.13 & 88.77 & 76.08 & 73.63 & 41.98 & 45.46 & 41.69 \\
\hline 
\end{tabular}
\begin{tablenotes}
    \footnotesize
    \item SCB: scene cue bank; w/o: without; w: with.
\end{tablenotes}
\end{threeparttable}
\end{table*}

\vspace{-0.1cm}
\subsubsection{Analysis on detecting ability} \label{subsec:Analysis on detecting ability}

For safety reasons in cooperative vehicle-infrastructure systems, it is more meaningful for detectors to tell THERE IS an object than the precise object location with tens of centimeter offsets. Thus, we conduct quantitative and qualitative analysis for the detecting ability. 

\noindent \textbf{Quantitative analysis.} We evaluate BEV AP and 3D AP for each category with 0.1 IoU. Overall, the detection performance of MOSE significantly surpasses BEVHight as shown in Table \ref{hete0.1}. Specifically, 3D AP of the car category improves by 5.05\%, 1.13\%, and 3.27\% in 'Easy', 'Moderate', and 'Hard' setting, respectively; 3D AP of the big vehicle improves by more than 10\% on average. This result reveals that MOSE has a stronger detection capability with much fewer undetected objects.

\begin{figure*}[t]
    \centering
    \includegraphics[width=0.8\textwidth]{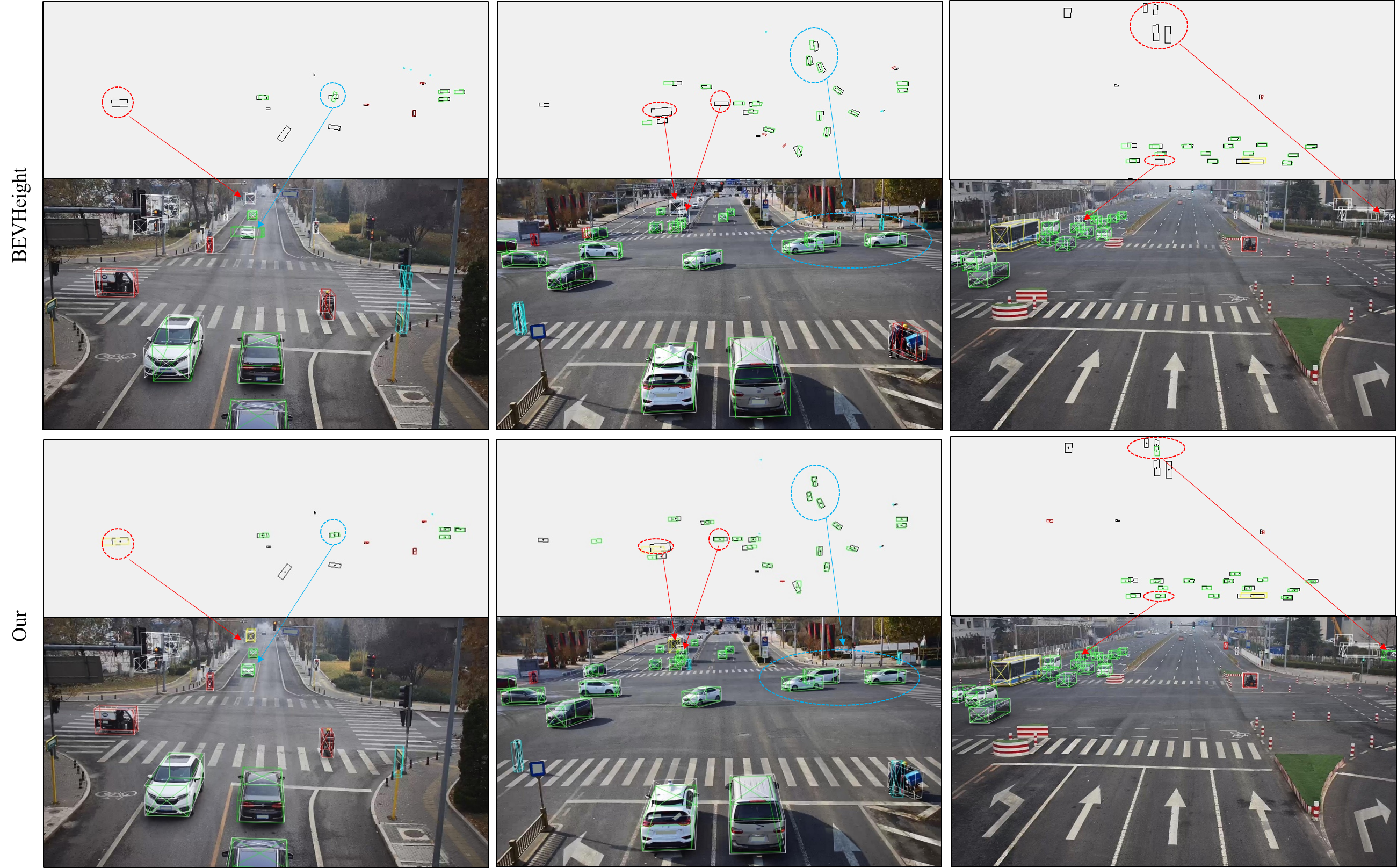}
    \vspace{-0.3cm}
    \caption{Qualitatively Comparisons on the Rope3D validation set in heterologous setting. In the BEV visualization, GT is the black box, while in the visual image, GT is white, and other colors are the detection results of different categories. }
    \label{fig:visualization_result}
    \vspace{-0.3cm}
\end{figure*}

\noindent \textbf{Qualitative analysis.} Fig.\ref{fig:visualization_result} shows the BEV and 3D detection results of MOSE and BEVHeight in three heterologous scenes. The objects directed by red arrows clearly represent MOSE reduces the undetected objects. The blue arrows show that MOSE obtains a more precise location than BEVHeight.

\begin{figure}
    \centering
    \includegraphics[width=0.48\textwidth]{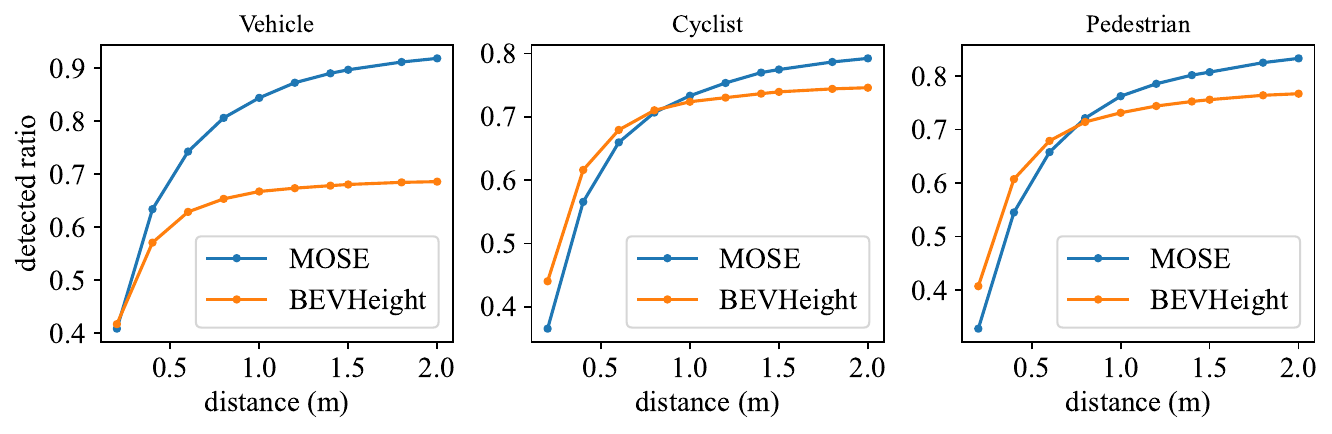}
    \vspace{-0.5cm}
    \caption{The detected ratio of predicted objects \emph{w.r.t.}, the GT number in different \emph{distance} thresholds. The \emph{distance} means Euclid distance of a ground-truth with its nearest predicted object.}
    \vspace{-0.2cm}
    \label{fig:dis_all}
\end{figure}

To further analyze the detecting ability, we calculate the distance between the ground-truth and predicted objects. By following the setting as in Table \ref{v2x} on the DAIR-V2X-I validation set, Fig.\ref{fig:dis_all} compares the ratio of detected objects \emph{w.r.t.}, the GT number in different distance thresholds.  In low distance threshold, our method is a little inferior to pedestrian and cyclist, which matches the results of pedestrian and cyclist in Table \ref{v2x}. Besides, the bounding boxes of pedestrian and cyclist categories are relatively small in the BEV, so little location errors will result in unstable 3D AP. With the increase of distance threshold, MOSE exceeds the BEVHeight, which means more objects are detected in our model overall. Specifically, our proposed method reflects high detecting ability when the distance threshold is more than 1m. For cooperative vehicle-infrastructure systems, the obstacle detected is taken more consideration and more meaningful than precision location. 

Note that our object queries are generated from a 2D detector. Different from 3D perception, a 2D detector takes no responsibility for the perception of the range of objects, which makes it relatively easy to improve 2D detection performance. Therefore, the 2D detector makes it easier to detect objects, \emph{i.e.}, obtain high recall, and then 2D lifting 3D helps our model achieve better 3D detecting capability.

\vspace{-0.1cm}
\subsection{Ablation Study}
The AP3D$|$R40 results on the Rope3D validation set in heterologous setting are reported for all ablation studies.
\vspace{-0.4cm}
\subsubsection{Analysis of scene cue bank}
\vspace{-0.1cm}
\noindent \textbf{Scene cue bank module.} Table \ref{ablation} shows the evaluation results of our model with or without scene cue bank. For both car and big vehicle categories, the model with scene cue bank has better detection results in all evaluation metrics. The ablation experiment results verify the proposed scene cue bank module is effective.

The 3D coordinates are determined by relative height and 2D pixel coordinates as shown in Fig.\ref{fig:2D_to_3D}. Because each position incorporates features from different objects with the same relative height in the scene cue bank, the richer features improve the accuracy of current relative height regression.


\noindent \textbf{Update Strategy.} We evaluate the impact of different strategies for updating the scene cue bank, including aggregating features from the full image, the object center points, and their 4 or 8 neighborhoods. As shown in Table \ref{update_strategy}, when the scene cue bank is updated by full-image features, the detection performance is even decreasing. The reason is that full-image features only add some useless ground information to the current location, which can not boost the detection performance. We find that the scene cues updated from the object center-point are effective, which inspires us to further conduct 4 and 8 neighborhood updating strategies. As shown in Table \ref{update_strategy}, utilizing 8-point neighborhood features produces the best results.

The above observations indicate the effectiveness of our proposed scene cue bank, and surrounding features of objects are helpful to the perception of relative height.

\begin{table}[h]
\small
\centering
\vspace{-0.1cm}
\caption{Analysis on different update strategies of scene cue bank.}\label{update_strategy}
\vspace{-0.3cm}
\begin{tabular}{c|p{1cm}<{\centering}|p{1cm}<{\centering}|p{1cm}<{\centering}}
\hline 
\multirow{2}{*}{update region} & \multicolumn{3}{c}{Car 3d AP (IoU=0.5)} \\
\cline{2-4}
~ & Easy & Mod. & Hard \\
\hline 
w/o SCB & 28.77 & 24.56 & 23.27\\
\hline 
full images & 21.57 & 17.96 & 17.68\\
 \hline
center-point & 29.82 & 24.79 & 23.42\\
 \hline 
4-point neighborhood & 29.31 & 25.06 & 23.68\\
 \hline
8-point neighborhood & \textbf{31.10} & \textbf{25.62} & \textbf{25.31}\\
\hline 
\end{tabular}
\end{table}

\subsubsection{Analysis of scene-based augmentation strategy}
In this section, we analyze the scene-based augmentation strategy. As described  in Sec. \ref{subsec:Implementation Details}, we increase the diversity of scenes by augmenting camera intrinsic and extrinsic parameters. Sec.\ref{subsec:Framework} introduces the scene-based augmentation related to frame number $\tau$. In this section, we conduct an ablation study about $\tau$ as shown in Table \ref{augmenation_strategy}. A smaller threshold results in insufficient scene cue information, while a larger will reduce the diversity of scenes at the specified epoch numbers. When the frame threshold is set to 1000-2000, our model obtains similar results.


\begin{table}[h]
\small
\centering
\caption{Analysis of scene-based augmentation strategies.}\label{augmenation_strategy}
\vspace{-0.3cm}
\begin{threeparttable}
\begin{tabular}{c|p{1cm}<{\centering}|p{1cm}<{\centering}|p{1cm}<{\centering}}
\hline 
\multirow{2}{*}{Frame threshold $\tau$} & \multicolumn{3}{c}{Car 3d AP (IoU=0.5)} \\
\cline{2-4}
~ & Easy & Mod. & Hard \\
\hline 
100 & 28.70 & 24.44 & 23.14 \\
\hline 
500 & 29.26 & 24.74 & 23.39 \\
 \hline
1000 & \textbf{31.10} & \textbf{25.62} & \textbf{25.31} \\
 \hline 
1500 & 29.21 & 24.92 & 23.53 \\
 \hline
2000 & 30.69 & 25.44 & 25.11 \\
\hline 
default & 27.60 & 23.96 & 22.70 \\
\hline 
\end{tabular}
\begin{tablenotes}
    \footnotesize
    \item default: The default frame number for each scene in the training set, i.e. no data augmentation.
\end{tablenotes}
\end{threeparttable}
\end{table}

\vspace{-0.1cm}
\section{Conclusion}
\vspace{-0.2cm}
\label{sec:conclusion}
In this paper, we propose a novel vision-based roadside 3D object detection framework, namely, MOSE. A scene cue bank is designed to aggregate scene cues from multiple frames of the same scene, which are indeed the frame-invariant and scene-specific features. The transformer-based decoder lifts the scene cues as well as incorporates the 3D position embeddings for 3D object location, which boosts generalization ability. With the 2D to 3D lifting strategy and scene-based augmentation strategy, MOSE ensures the high detecting ability and locating precision of vehicles in heterologous scenes. The extensive experiment results on two public benchmarks demonstrate that our method achieves state-of-the-art performance for detecting 3D objects on roadside cameras. We hope this work will pave a new way for concentrating on the frame-invariant features aimed at the images from roadside cameras.

{\small
\bibliographystyle{ieee_fullname}
\bibliography{egbib}
}

\clearpage
\setcounter{page}{1}
\maketitlesupplementary

\section{Appendix}

\subsection{Analysis of Inference}
\vspace{-0.1cm}
\noindent \textbf{Algorithm} In the training process, the scene cue bank is updated in a momentum way based on scene-based augmentation strategy as shown in Alg.\ref{training_algorithm}. In the inference process, we first sample a subset from the testing set to update the scene cue bank by average scene cues, and then conduct the 3D detection for the target frame using the memorized bank, as shown in Alg.\ref{testing_algorithm}. In general, our method is deployed to the infrastructure with the specific scene, so its inference process is scene by scene. For a specific scene, the images collected in the early stage can be used to update the scene cue bank, which is convenient and does not require updating bank every time the target frame is detected.

\noindent \textbf{Costing and memory} The first loop is only conducted once per scene, so the inference time of a frame is approximate to the method without the scene bank. For one scene, the dimensions of the bank is $1\times\frac{H}{8}\times \frac{W}{8}\times d$, where $H$, $W$, and $d$ is 1024, 1536, and 256, respectively. Therefore, the extra memory of MOSE is only 6.3M.

\vspace{-0.3cm}
\begin{algorithm}[h]
\small
\caption{Algorithm for Inference} \label{testing_algorithm}
\SetAlgoLined
\KwIn{testing set $D^{te}$ with $S$ samples, \\
scene cue bank $\mathcal{U} \in \mathbb{R}^{S\times\frac{H}{8}\times \frac{W}{8}\times d}$, \\
object counter $\mathcal{N} \in \mathbb{R}^{S\times\frac{H}{8}\times \frac{W}{8}}$ of each pixel}
\KwOut{3D bounding bbox $B$}
initialize: $\mathcal{U}$, $\mathcal{N}$ \\
Sample a subset $D^{te}_{sub}$ with $S_{sub}$ samples from $D^{te}$ \\
\For{$b\leftarrow 1$ \KwTo $S_{sub}$}{
Sample a sample $I_t^s$ from $D^{te}_{sub}$ \\
Extract features ${\bm f}^s$ of current frame, memorized $\hat{\bm{f}}_{t-1}^s$ \\
Obtain scene cues mask $\bm{M}^s$ \\
Update $\hat{\bm{f}}_t^s \leftarrow \hat{\bm{f}}_{t-1}^s+[\frac{\mathcal{N}_i-1}{\mathcal{N}_i}*\hat{\bm{f}}_{t-1}^s+\frac{{\bm f}^s}{\mathcal{N}_i}]$ \\
Update $\mathcal{N}_t^s \leftarrow \mathcal{N}_{t-1}^s + \bm{M}^s$ 
}
\For{$b\leftarrow 1$ \KwTo $S$}{
Sample a sample $I_t^s$ from $D^{te}$ \\
Obtain memorized $\hat{\bm{f}^s}$ of scene s \\
Obtain output $B_i={\cal F}(I_t^s, \hat{\bm{f}^s})$ \\
}
\end{algorithm}
\vspace{-0.5cm}

\subsection{Analysis of the decoder blocks}
\vspace{-0.1cm}
As shown in Table \ref{decoder_blocks}, we also ablate the configuration of decoder layers in the 3D detector. With the number of decoder block layer  increasing, the performance of the model improves. After the decoder block reaches 6 layers, the performance gradually stabilizes, so our model uses 6 decoder blocks.

\begin{table}[h]
\small
\centering
\caption{Analysis of the decoder block numbers.}\label{decoder_blocks}
\vspace{-0.3cm}
\begin{tabular}{c|p{1cm}<{\centering}|p{1cm}<{\centering}|p{1cm}<{\centering}}
\hline 
\multirow{2}{*}{decoder block setting} & \multicolumn{3}{c}{Car 3D AP (IoU=0.5)} \\
\cline{2-4}
~ & Easy & Mod. & Hard \\
\hline 
3 & 27.96 & 24.04 & 22.74 \\
\hline 
4 & 28.02 & 24.17 & 22.84 \\
 \hline
5 & 30.83 & 25.43 & 25.10 \\
 \hline 
6 & \textbf{31.10} & 25.62 & \textbf{25.31} \\
 \hline
7 & 30.81 & \textbf{26.64} & 25.27 \\
\hline 
8 & 30.80 & 25.43 & 25.10 \\
\hline 
\end{tabular}
\vspace{-0.3cm}
\end{table}

\subsection{Analysis of the distance error}
\vspace{-0.1cm}
To further evaluate the detection ability of our method, We divide the distance between 0-200m into four levels, with each level of 50m, and calculate the distance error between detected objects and the matched ground-truth. The matching method is based on the IOU in the pixel coordinate system. Note that all objects including the occluded, truncated, and small objects are calculated. The distance error($E$) is defined as:

\begin{equation} 
    E = |d_p-d_g| / d_g \times 100 \%
\end{equation}
where $d_p$ and $d_g$ are the depth of the prediction and ground-truth, respectively.

Table \ref{table:distance_error_rope3d} and Table \ref{table:distance_error_v2x} show the detected numbers and the distance errors of BEVHeight \cite{BEVheight} and MOSE on the Rope3D \cite{Rope3D} validation set in heterologous settings and on the DAIR-V2X \cite{DAIR-V2X} validation set in homologous settings, respectively.

On the Rope3D dataset, higher TP in Table \ref{table:distance_error_rope3d} shows that MOSE has better performance in detection ability, which is crucial for practical applications. Besides, MOSE achieves more precise locations, especially for long-distance (\emph{i.e.}, $>100m$) objects. 

On the DAIR-V2X dataset, Table \ref{table:distance_error_v2x} quantifies that MOSE significantly reduces undetected objects, demonstrating a strong detection ability for the challenging cyclist and pedestrian categories. Similarly, for long-distance objects, MOSE outperforms BEVHeight in all categories. 

Overall, MOSE obtains higher recall and lower distance errors and has tremendous superiority in long-distance object location.

\begin{table*}[h]
\small
\centering
\caption{Analysis of the distance error on the Rope3D validation set in heterologous settings.}\label{table:distance_error_rope3d}
\vspace{-0.1cm}
\begin{threeparttable}
\begin{tabular}{c|c|c|c|c|c|c|c|c}
\hline 
\multirow{2}{*}{Category} & \multirow{2}{*}{Method} & \multirow{2}{*}{GT} & \multirow{2}{*}{TP} & \multirow{2}{*}{Recall} & \multicolumn{4}{c}{distance error $E$ (\%)} \\
\cline{6-9}
~ & ~ & ~ & ~ & ~ & [0,50] & [50,100] & [100,150] & [150,200] \\
\hline 
\multirow{2}{*}{Car} & BEVHeight & \multirow{2}{*}{37810} & 28912 & 76.47 & 2.46 & 2.05 & 4.48 & 10.88 \\
~ & MOSE & ~ & 32431 & \underline{85.77} & 2.08 & 1.70 & \underline{3.82} & \underline{7.40} \\
\hline 
\multirow{2}{*}{Big Vehicle} & BEVHeight & \multirow{2}{*}{5613} & 1556 & 27.72 & 4.75 & 2.09 & 4.00 & 3.48 \\
~ & MOSE & ~ & 3143 & \underline{56.00} & 5.18 & 5.66 & \underline{3.84} & \underline{3.45} \\
\hline 
\end{tabular}
\begin{tablenotes}
    \footnotesize
    \item GT : the bounding box number of ground truth.\\
          TP : the matched box number between prediction results and ground-truth.
\end{tablenotes}
\end{threeparttable}
\end{table*}

\begin{table*}[h]
\small
\centering
\caption{Analysis of the distance error on the DAIR-V2X validation set in homologous settings.}\label{table:distance_error_v2x}
\vspace{-0.1cm}
\begin{threeparttable}
\begin{tabular}{c|c|c|c|c|c|c|c|c}
\hline 
\multirow{2}{*}{Category} & \multirow{2}{*}{Method} & \multirow{2}{*}{GT} & \multirow{2}{*}{TP} & \multirow{2}{*}{Recall} & \multicolumn{4}{c}{distance error $E$ (\%)} \\
\cline{6-9}
~ & ~ & ~ & ~ & ~ & [0,50] & [50,100] & [100,150] & [150,200] \\
\hline 
\multirow{2}{*}{Vehicle} & BEVHeight & \multirow{2}{*}{30133} & 20913 & 69.40 & 0.76 & 0.91 & 9.50 & 40.67 \\
~ & MOSE & ~ & 29196 & \underline{96.89} & 0.76 & 0.64 & \underline{0.92} & \underline{2.12} \\
\hline 
\multirow{2}{*}{Cyclist} & BEVHeight & \multirow{2}{*}{10196} & 7724 & 75.76 & 0.52 & 0.74 & 1.62 & - \\
~ & MOSE & ~ & 8436 & \underline{82.74} & 0.62 & 0.90 & \underline{1.47} & - \\
\hline 
\multirow{2}{*}{Pedestrian} & BEVHeight & \multirow{2}{*}{7344} & 5673 & 77.25 & 0.70 & 0.65 & 2.51 & - \\
~ & MOSE & ~ & 6338 & \underline{86.30} & 0.77 & 0.82 & \underline{1.26} & - \\
\hline 
\end{tabular}
\begin{tablenotes}
    \footnotesize
    \item GT : the bounding box number of ground truth.\\
          TP : the matched box number between prediction results and ground-truth. \\
          \hspace{0.1cm} - : There are not matched objects in this distance range. 
\end{tablenotes}
\end{threeparttable}
\end{table*}

\vspace{-0.1cm}
\subsection{More Visualizations}
\begin{figure*}[t]
    \centering
    \includegraphics[width=0.9\textwidth]{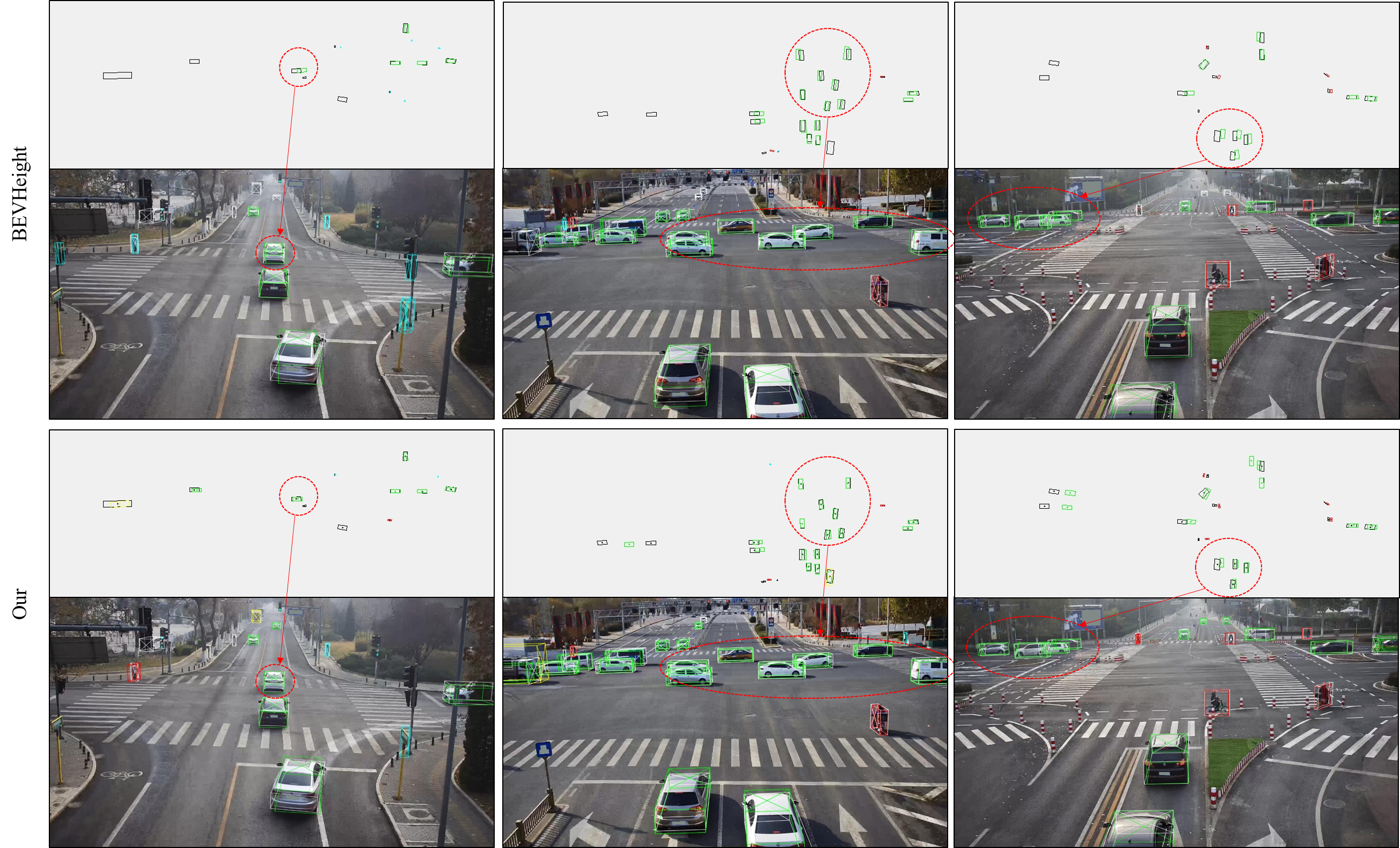}
    \caption{Visualization results of our proposed MOSE and BEVHeight on the Rope3D validation set in heterologous settings.In the BEV visualization, GT is the black box, while in the visual image, GT is white, and other colors are the detection results of different categories. The circled objects represent MOSE obtains more precise locations.}
    \label{fig:vis_app}
\end{figure*}

\vspace{-0.1cm}
Fig.\ref{fig:vis_app} shows more qualitative comparisons between our proposed MOSE and BEVHeight on the Rope3D validation set in heterologous settings. As shown in the BEV visualization, it is clear that MOSE obtains more precise locations, especially for the occluded and long-distance objects.

\end{document}